\pdfoutput=1
\documentclass[10pt,twocolumn,letterpaper]{article}

\usepackage{cvpr}
\usepackage{times}
\usepackage{epsfig}
\usepackage{graphicx}
\usepackage{amsmath}
\usepackage{amssymb}

\usepackage[breaklinks=true,bookmarks=false,pagebackref=true,colorlinks]{hyperref}
\cvprfinalcopy 

\ifcvprfinal\fi
\begin{document}

\title{Deep Cuboid Detection: Beyond 2D Bounding Boxes}

\author{\large Debidatta Dwibedi\thanks{Work done as an intern at Magic Leap, Inc. while a student at
Carnegie Mellon University}\\
\normalsize Carnegie Mellon University\\
{\tt\small debidatta@cmu.edu}
\and
\large Tomasz Malisiewicz \quad   Vijay Badrinarayanan \quad Andrew Rabinovich\\
\normalsize Magic Leap, Inc.\\
{\tt\small {\{tmalisiewicz,vbadrinarayanan,arabinovich\}@magicleap.com}}}

\maketitle


\begin{abstract}

We present a Deep Cuboid Detector which takes a consumer-quality RGB image of a cluttered scene and localizes all 3D cuboids (box-like objects). Contrary to classical approaches which fit a 3D model from low-level cues like corners, edges, and vanishing points, we propose an end-to-end deep learning system to detect cuboids across many semantic categories (e.g., ovens, shipping boxes, and furniture). We localize cuboids with a 2D bounding box, and simultaneously localize the cuboid's corners, effectively producing a 3D interpretation of box-like objects. We refine keypoints by pooling convolutional features iteratively, improving the baseline method significantly. Our deep learning cuboid detector is trained in an end-to-end fashion and is suitable for real-time applications in augmented reality (AR) and robotics. 

\end{abstract}

\section{Introduction}
Building a 3D representation of the world from a single monocular image is an important problem in computer vision. In some applications, we have the advantage of having explicit 3D models and try to localize these objects in the world while estimating their pose. But without such 3D models, we must reason about the world in terms of simple combinations of geometric shapes like cuboids, cylinders, and spheres. Such primitives, referred to as \emph{geons} by Biederman \cite{biederman1987recognition}, are easy for humans to reason about. Humans can effortlessly make coarse estimates about the pose of these simple geometric primitives and even compare geometric parameters like length, radius or area across disparate instances. While many objects are composed of multiple geometric primitives, a large number of real objects can be well approximated by as little as one primitive.

\begin{figure}[t]
\begin{center}
  \includegraphics[width=\linewidth]{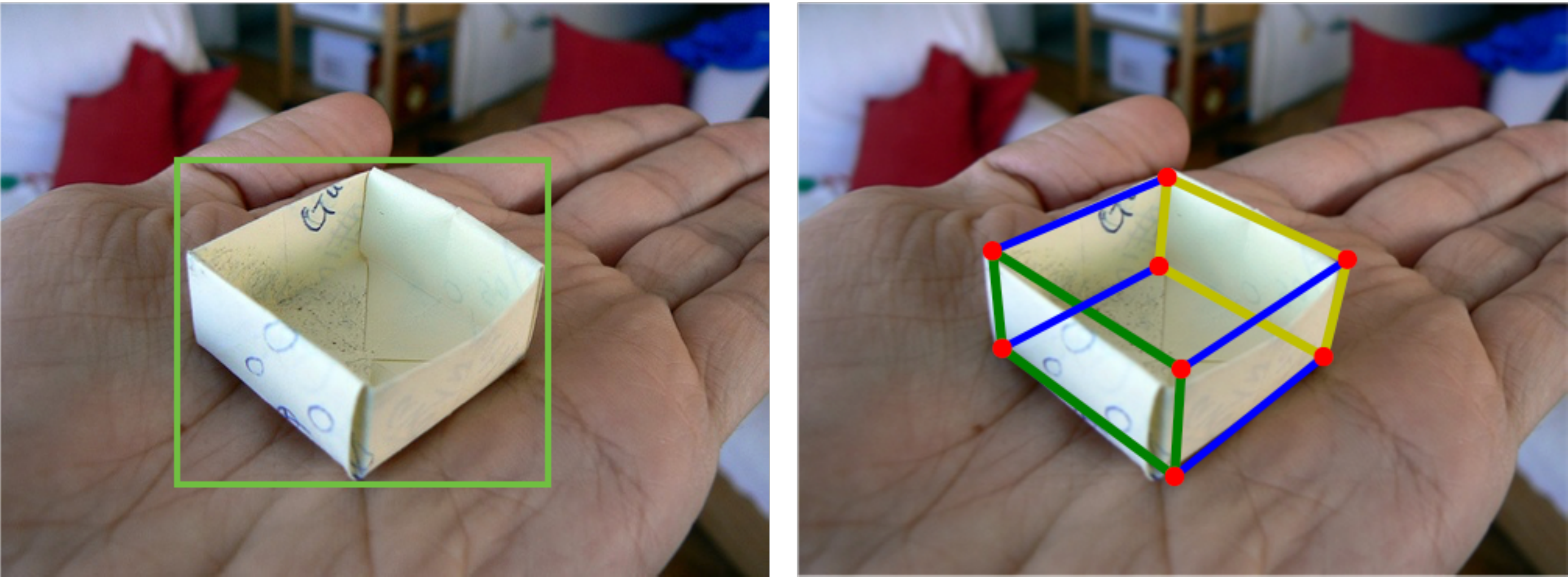}
\end{center}
  \caption{{\bf 2D Object detection vs. 3D Cuboid detection.} Our task is to find all cuboids inside a monocular image and localize their vertices. Our deep learning system is trained in an end-to-end fashion, runs in real-time, and works on RGB images of cluttered scenes captured using a consumer-grade camera.}
\label{fig:teaser}
\end{figure}

For example, let us consider a common shape that we see everyday: \emph{the box}. Many everyday objects can geometrically be classified as a box (\eg., shipping boxes, cabinets, washing machines, dice, microwaves, desktop computers). Boxes (or cuboids, as we call them in this paper) span a diverse set of everyday object instances, yet it is very easy for humans to fit an imaginary cuboid to these objects and in doing so they localize its vertices and faces. People can also compare the dimensions of different box-like objects even though they are not aware of the exact dimensions or even if the object is not a perfect cuboid. In this work, we specifically deal with the problem of detecting class agnostic geometric entities like cuboids. By class agnostic, we mean that we do not differentiate between a shipping box, a microwave oven, or a cabinet. All of these boxy objects are represented with the same simplified concept called \emph{cuboid} in the rest of the paper. 

\begin{figure*}[t]
\begin{center}
  \includegraphics[width=\linewidth]{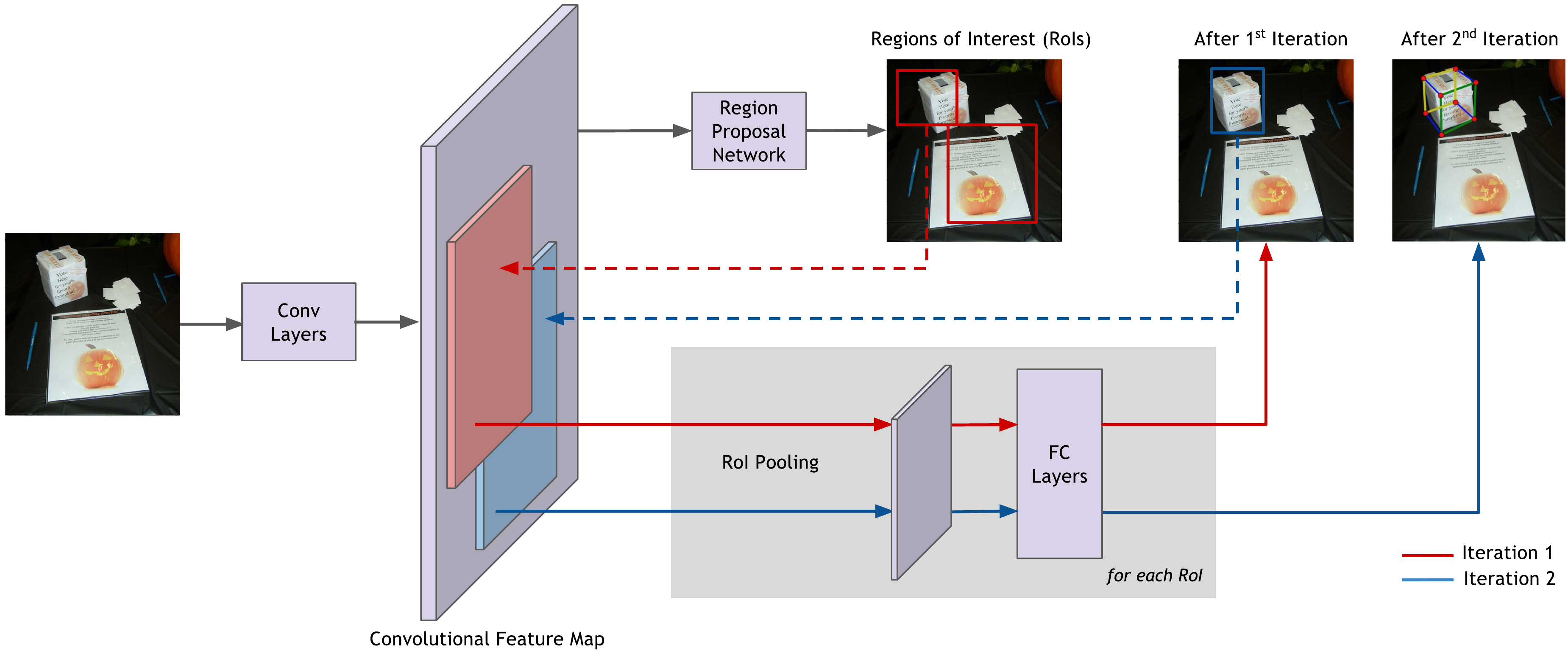}
\end{center}
  \caption{{\bf Deep Cuboid Detection Pipeline.} The first step is to find Regions of Interest (RoIs) in the image where a cuboid might be present. We train a Region Proposal Network (RPN) to output such regions. Then, features corresponding to each RoI are pooled from a convolutional feature map (\eg., \textit{conv5} in VGG-M). These pooled features are passed though two fully connected layers just like the region-based classification network in Faster R-CNN (see model details in Section~\ref{sec:model}). However, instead of just producing a 2D bounding box, we also output the normalized offsets of the vertices from the center of the region. Finally, we refine our predictions by performing iterative feature pooling (see Section~\ref{sec:ifp}). The dashed lines show the regions from which features will be pooled.}
\label{fig:network}

\end{figure*}

Detecting boxy objects in images and extracting 3D information like pose helps overall scene understanding. Many high-level semantic problems have been tackled by first detecting boxes in a scene (\eg., extracting the free space in a room by reducing the objects in a scene to boxes~\cite{gupta20113d}, estimating the support surfaces in the scene~\cite{jia20133d,gupta2010blocks} and estimating the scene layout~\cite{geiger2011joint}).

A perfect cuboid detector opens up a plethora of possibilities for augmented reality (AR), human-computer interaction (HCI), autonomous vehicles, drones, and robotics in general. The cuboid detector can be used as follows:
\begin{itemize}

\item For Augmented Reality, cuboid vertex localization followed by 6-dof pose estimation allows a content creator to use the cuboid-centric coordinate system defined by a stationary box to drive character animation. Because we also know the volume of space occupied by the stationary cuboid, animated characters can jump on the box, hide behind it, and even start drawing on one of the box's faces.
\item For Human-Computer Interaction, a hand-held cuboid can be used as a lightweight game controller. A stationary camera observing the hand-held cube is able to estimate the cube's pose, effectively tracking the cube in 3D space. In essence, the cuboid can serve as a means to improve interaction in AR systems(\eg., the tabletop AR demo using cuboids~\cite{zheng2012interactive}).
\item For autonomous vehicles, 3D cuboid detection allows the vehicle to reason about the spatial extent of rare objects that might be missing in supervised training set. By reasoning about the pose of objects in a class-agnostic manner, autonomous vehicles can be safer drivers.
\item For drones, man-made structures such as buildings and houses can be well-approximated with cuboids, assisting navigation during unmanned flights.
\item For robotics in general, detecting boxy objects in images and extracting their 3D information like pose helps overall scene understanding. Placing a handful of cuboids in a scene (instead of Aruco markers) can making pose tracking more robust for SLAM applications.
\end{itemize}

In general, one can formulate the 3D object detection problem as follows: fit a 3D bounding box to objects in an RGB-D image~\cite{song2014sliding,song2015deep,jia20133d,shao2014imagining}, detect 3D keypoints in an RGB image~\cite{tulsiani2015viewpoints,wu2016single}, or perform 3D model to 2D image alignment~\cite{su2015render,gupta2015inferring,bansal2016marr}. In this paper, we follow the keypoint-based formulation. Because an image might contain multiple cuboids as well as lots of clutter, we must first procure a shortlist of regions of interest that correspond to cuboids. In addition to the 2D bounding box enclosing the cuboid, we estimate the location of all $8$ vertices. 

\subsection{From Object Detection to Cuboid Detection}

Deep learning~\cite{krizhevsky2012imagenet} has revolutionized image recognition in the past few years. Many state-of-the-art methods in object detection today are built on top of deep networks that were trained for the task for image classification. Not only has the accuracy of object detection increased, there are many approaches that also run in real-time~\cite{girshick2014rich,ren2015faster,liu2015ssd}. We take advantage of these advances to detect boxy objects in a scene. Unlike the usual task of object detection, we want more than the bounding box of the object. Instead, we want to localize the vertices of the cuboids (see Figure~\ref{fig:teaser}). Another aspect where we differ from the object detection task is that we do not care about the class of the cuboids that we are detecting. For our purpose, we deal with only two classes: cuboid and not-a-cuboid.

Even if a cuboid is a geometric object that can be easily parameterized, we advocate the use of deep learning to detect them in scenes. Classically, one approach to detect cuboids is to detect the edges and try to fit the model of a cuboid to these edges. Hence, robust edge selection is an important step. However, this becomes difficult when there are misleading textures on cuboidal surfaces, if edges and corners are occluded or the scene contains considerable background clutter. It is a difficult task to classify whether a given line belongs to a given cuboid with purely local features. Hence, we propose to detect cuboids the same way we detect classes like cars and aeroplanes in images, using a data-driven approach. However, the task of fitting a single label 'cuboid' to boxy objects in the scene is not trivial as the label is spread over many categories like houses, washing machines, ballot boxes, etc. In our experiments, we show how a CNN is successful to learn features that help us identify cuboids in different scenes.

\subsection{Our Contributions}
Our main contribution is a deep learning model that jointly performs cuboid detection and vertex localization. We explore various ways to improve the detection and localization accuracy produced by a baseline CNN. The key idea is to first detect the object of interest and then make coarse predictions regarding the location of its vertices. The next stage acts as an attention mechanism, performing refinement of vertices by only looking at regions with high probability of being a cuboid. We improve localization accuracy with an iterative feature pooling mechanism (see Section~\ref{sec:ifp}), study the effect of combining cuboid-related losses( see Section~\ref{sec:multi-task}), experiment with alternate parametrizations (see Section~\ref{sec:parametrizations}), and study the effect of different size base networks (see Section~\ref{sec:training-set-experiment}).

\section{Related Work}

Classical ideas on 3D scene and object recognition originate in Robert's Blocks-World~\cite{roberts1963machine} and Biederman's Recognition-by-Components~\cite{biederman1987recognition}. These early works were overly reliant on bottom-up image processing, thus never worked satisfactorily on real images. Many modern approaches utilize a large training database of 3D models and some kind of learning for 2d-to-3d alignment~\cite{aubry2014seeing,bansal2016marr,wu2016single,choy20163d}.

Cuboid detection has also been approached with geometry-based methods~\cite{wilczkowiak2005using,hedau2012recovering,lee2009geometric,gupta20113d}. Shortly after the success of Deformable Parts Model, researchers extended HOG-based models to cuboids~\cite{xiao2012localizing,fidler20123d}. RGB-D based approaches to cuboid detection are also common~\cite{jiang2013linear}.

Our work revisits the problem of cuboid detection that Xiao \etal. introduced in \cite{xiao2012localizing}. They use a Deformable Parts-based Model that uses HOG classifiers to detect cuboid vertices in different views. Their model has four components for scoring a final cuboid configuration: score from the HOG classifier, 2D vertex displacement, edge alignment score and a 3D shape score that takes into account how close the predicted vertices are to a cuboid in 3D. Their approach jointly optimizes over visual evidence (corners and edges) found in the image while penalizing the predictions that stray too far from an actual 3D cuboid. A major limitation of their approach is the computationally expensive test-time iterative optimization step. Not only is their HOG-based model inferior to its modern CNN-based counterpart (as we demonstrate in our experiments), their approach takes more than a minute to process a single image.

Today the best methods in object detection owe their performance to convolutional neural networks. Krizhevsky \etal.~\cite{krizhevsky2012imagenet} showed that CNNs were superior to existing methods for the task of image classification. To localize an object in an image, the image is broken down into regions and these regions are classified instead. This approach was first shown to work in \cite{girshick2014rich}. This was followed by attempts to make object detection work almost real-time~\cite{girshick2015fast,ren2015faster}. More recently, researchers have been able to make object detection faster by performing detection in a single step~\cite{liu2015ssd,redmon2015you}. These approaches run at about 50-60 frames per second which opens up the possibility for adapting such network architectures to real-time cuboid localization.

3D object localization using keypoints is commonly studied~\cite{savarese20073d,hoiem2011representations,hsiao2010making,collet2011moped,lim2014fpm,crivellaro2015novel}. 3D keypoint detection using deep networks is also gaining popularity~\cite{wu2016single, tulsiani2015viewpoints}. There has been a resurgence in work that aims to align 3D models of objects in single view images~\cite{gupta2015aligning,kar2015category,xiang2015data}.

The iterative vertex refinement component of our approach is similar to the iterative error feedback approach of \cite{carreira2015human}, the network cascades in \cite{dai2015instance}, the iterative bounding box regression of Multi-region CNN~\cite{gidaris2015object} and Inside-Outside Networks~\cite{bell2015inside}. Such iterative models have been reinterpreted as Recurrent Neural Networks in \cite{belagiannis2016recurrent,bulat2016human}, and while most applications focus on human pose estimation, the ideas can easily be extended to cuboid detection.

\begin{figure}[t]
\begin{center}
  \includegraphics[width=0.6\linewidth]{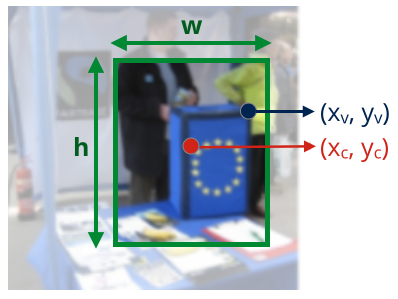}
\end{center}
  \caption{{\bf RoI-Normalized Coordinates} Vertices are represented as offsets from the center of the RoI and normalized by the region's width/height. $(x_v,y_v)$ is a vertex and $(x_c,y_c)$ is the center.}
\label{fig:gt}
\end{figure}

\begin{figure*}[ht]
\begin{center}
  \includegraphics[width=0.9\linewidth]{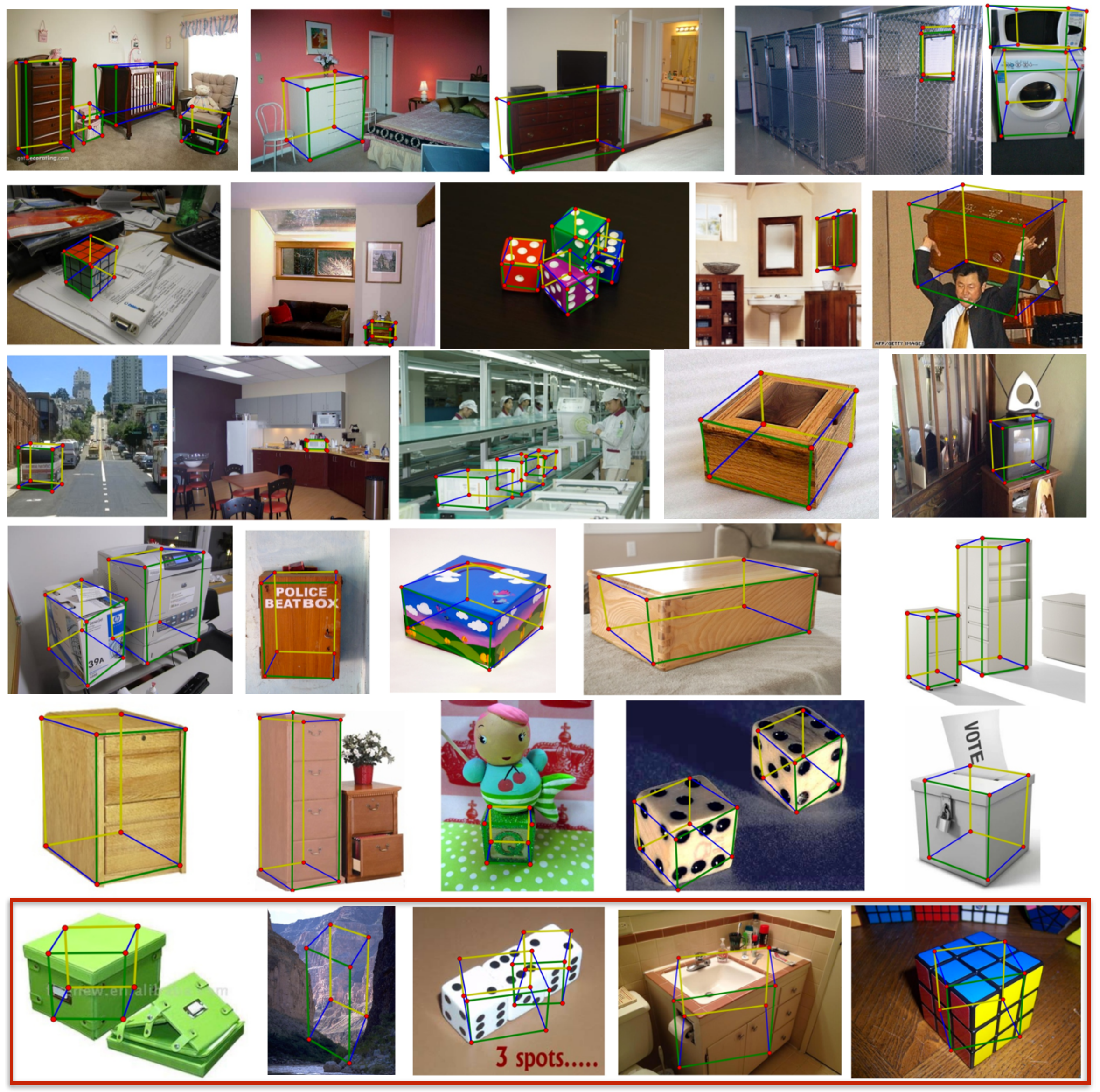}
\end{center}
  \caption{{\bf Deep Cuboid Detections using VGG16 + Iterative.} Our system is able to localize the vertices of cuboids in consumer-grade RGB images. We can handle both objects like boxes (that are perfectly modeled by a cuboid) as well as objects like sinks (that are only approximate cuboids). Last row: failures of our system (see \ref{sec:parametrizations} for more details).}
\label{fig:example-detections}
\end{figure*}

\section{Deep Cuboid Detection}
\label{section:dcd}
We define any geometric shape as a tuple of N parameters. These parameters might be geometric in nature like the radius of a sphere or the length, width and height of a cuboid. However, a more general way to parameterize any geometric primitive is to represent it as a collection of points on the surface of the primitive. If we choose a random point on the surface of the primitive, it might not necessarily be localizable from a computer-vision point of view. Ideally, we want the set of parametrization points to be \emph{geometrically informative} and \emph{visually discriminative}~\cite{fouhey2013data}. Specifically, in the case of cuboids, these special keypoints are the same as the cuboid's corners.\footnote{When discussing cuboids, we use the terms vertex, corner, and keypoint, interchangeably.} 

We define a cuboid as a tuple of $8$ vertices where each point is denoted by its coordinates $(x,y)$ in the image. In our default parametrization, we use the $16$ coordinate parameters to represent a cuboid in an image. This, however, is an overparameterization as $16$ parameters are not required to represent a cuboid (see Section~\ref{sec:parametrizations} for a discussion on alternate cuboid parametrizations).

\subsection{Network Architecture and Loss Functions}
\label{sec:model}
Our network is made up of several key components: the CNN Tower, the Region Proposal Network (RPN), R-CNN classifier and regressor, and a novel iterative feature pooling procedure which refines cuboid vertex estimates. Our architecture is visually depicted in Figure~\ref{fig:network}.

\begin{enumerate}
    \item The CNN Tower is the pre-trained fully convolutional part of popular ConvNets like VGG and ResNets. We refer to the last layer of the tower as the convolutional feature map (e.g., conv5 in VGG16 is of size $M \times N \times 512$).
    \item The RPN is a fully convolutional network that maps every cell in the feature map to a distribution over K multi-scale anchor-boxes, bounding box offsets, and objectness scores. The RPN has two associated losses: log loss for objectness and smooth L1 for bounding box regression. Our RPN uses 512 $3 \times 3$ filters, then 18  $1\times1$ filters for objectness and 36  $1\times1$ filters for bounding box offsets. (See Faster R-CNN~\cite{ren2015faster} for more details.)
    \item The RoI pooling layer uses max pooling to convert the features inside any valid region of interest into a small fixed-size feature map. E.g., for conv5 of size $M \times N \times 512$, the pooling layer produces an output of size $7 \times 7 \times 512$, independent of the input region’s aspect ratio and scale. RoI pooling is the non-hierarchical version of SPPNet~\cite{he2014spatial}.
    \item The RCNN regressor is then applied to each fixed-size feature vector, outputting a cuboidness score, bounding box offsets ($4$ numbers), and 8 cuboid vertex locations ($16$ numbers). The bounding box regression values  ($\Delta x$, $\Delta y$, $\Delta w$, $\Delta h$) \cite{girshick2015fast} are used to fit the initial object proposal tightly around the object. The vertex locations are encoded as offsets from the center of the RoI and are normalized by the proposal width/height as shown in Figure~\ref{fig:gt}. The ground truth targets for each vertex are:
\begin{align*}
t_x = \frac{x_v - x_c}{w} &, t_y = \frac{y_v - y_c}{h}\\
\end{align*} The RCNN uses two fully connected layers ($4096$ neurons each), and has three associated losses: log loss for cuboidness and smooth $L_1$ for both bounding box and vertex regression.
    \item When viewed in unison, the RoI pooling and RCNN layers act as a refinement mechanism, mapping an input box to an improved one, given the feature map. This allows us to apply the last part of our network multiple times --- a step we call iterative feature pooling(See section ~\ref{sec:ifp}).
\end{enumerate}

The loss function used in the RPN consists of $L_{anchor-cls}$, the log loss over two classes (cuboid vs. not cuboid) and $L_{anchor-reg}$, the Smooth $L_1$ loss~\cite{girshick2015fast} of the bounding box regression values for each anchor box. The loss function for the R-CNN is made up of $L_{ROI-cls}$, the log loss over two classes (cuboid vs. not cuboid), $L_{ROI-reg}$, the Smooth $L_1$ loss of the bounding box regression values for the RoI and $L_{ROI-corner}$, the Smooth $L_1$ loss over the RoI's predicted vertex locations. We also refer to the last term as the corner regression loss. In the experiments section, we report how adding these additional tasks affects the detector performance (see Table~\ref{table:incremental}).  

The complete loss function is a weighted sum of the above mentioned losses and can be written as follows:
\begin{align*}
    L =  &\lambda_1 L_{anchor-cls} + \lambda_2 L_{anchor-reg} + \\
    &\lambda_3 L_{ROI-cls} + \lambda_4 L_{ROI-reg} +  \lambda_5 L_{ROI-corner}
\end{align*}

We use Caffe~\cite{jia2014caffe} for our experiments and build on top of the implementation of Faster R-CNN by Girshick \etal.~\cite{girshick2015fast}. For all experiments we start with either the VGG-M~\cite{chatfield2014return} or VGG16~\cite{simonyan2014very} networks that have been pre-trained for the task of image classification on ImageNet. VGG-M is a smaller model with $7$ layers while VGG16 contains $16$ layers. We fine-tune all models for $50$K iterations using SGD with a learning rate of $0.001$ and reduce the learning rate by a factor of $10$ after $30$K iterations. We also use a momentum of $0.9$, weight decay of $0.0005$ and dropout of $0.5$. We do not perform any stage-wise training. Instead, we jointly optimize over all tasks and have kept the value of all the loss weights as one in our experiments (\ie., $\lambda_i = 1$).

\section{Data}

We use the SUN Primitive dataset~\cite{xiao2012localizing} to train our deep cuboid detector. This is the largest publicly available cuboid dataset --- it consists of $3516$ images and is a mix of indoor scenes with lots of clutter, internet images containing only a single cuboid and outdoor images of buildings that also look like cuboids. Both cuboid bounding boxes and cuboid vertices have ground-truth annotations. In this dataset, we have $785$ images with $1269$ annotated cuboids. The rest of the images are negatives --- they do not contain any cuboids. We make a split to create a training set of $3000$ images and $516$ test images. We augment the $3000$ images with their horizontally flipped versions while training, but cuboid-specific data augmentation~\cite{hejrati2016categorizing} strategies are likely to further improve performance.

\section{Experiments}
\label{section:experiments}
\begin{figure*}
\begin{center}
\begin{tabular}{cc}
  \includegraphics[width=0.5\textwidth]{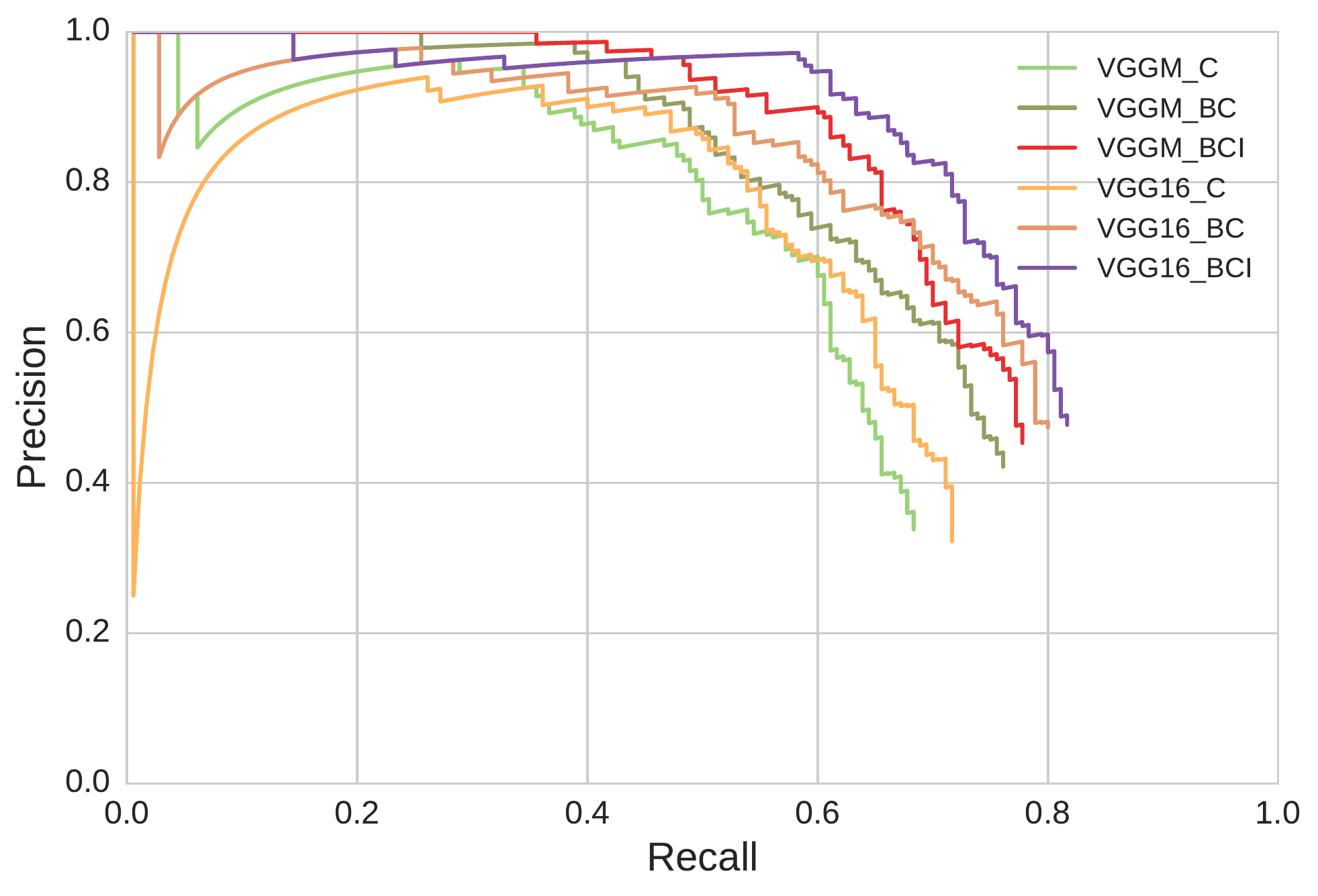} &   \includegraphics[width=0.5\textwidth]{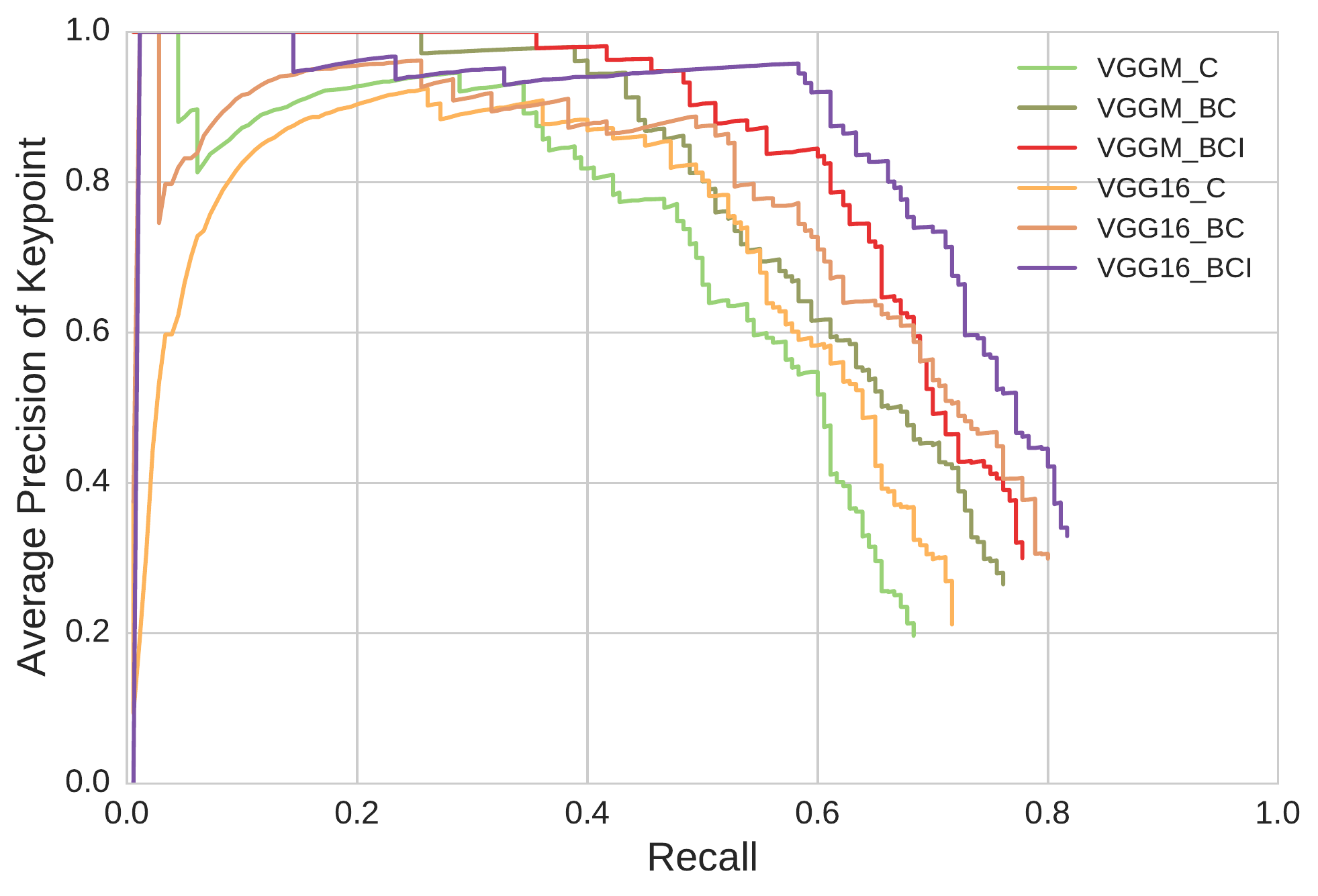} \\
(a) Precision-Recall Curve & (b) APK vs. Recall \\[6pt]
 \includegraphics[width=0.5\textwidth]{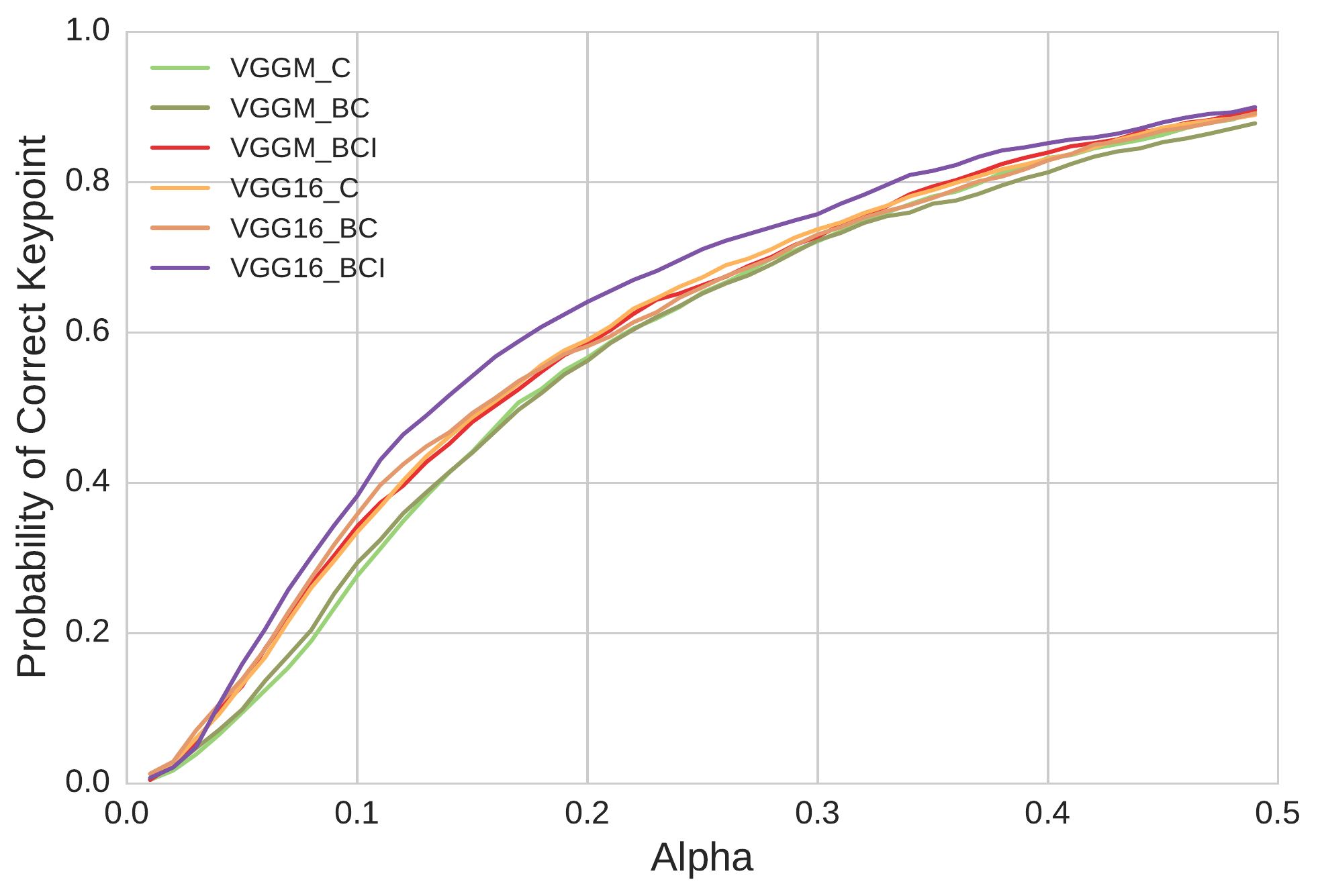} &   \includegraphics[width=0.5\textwidth]{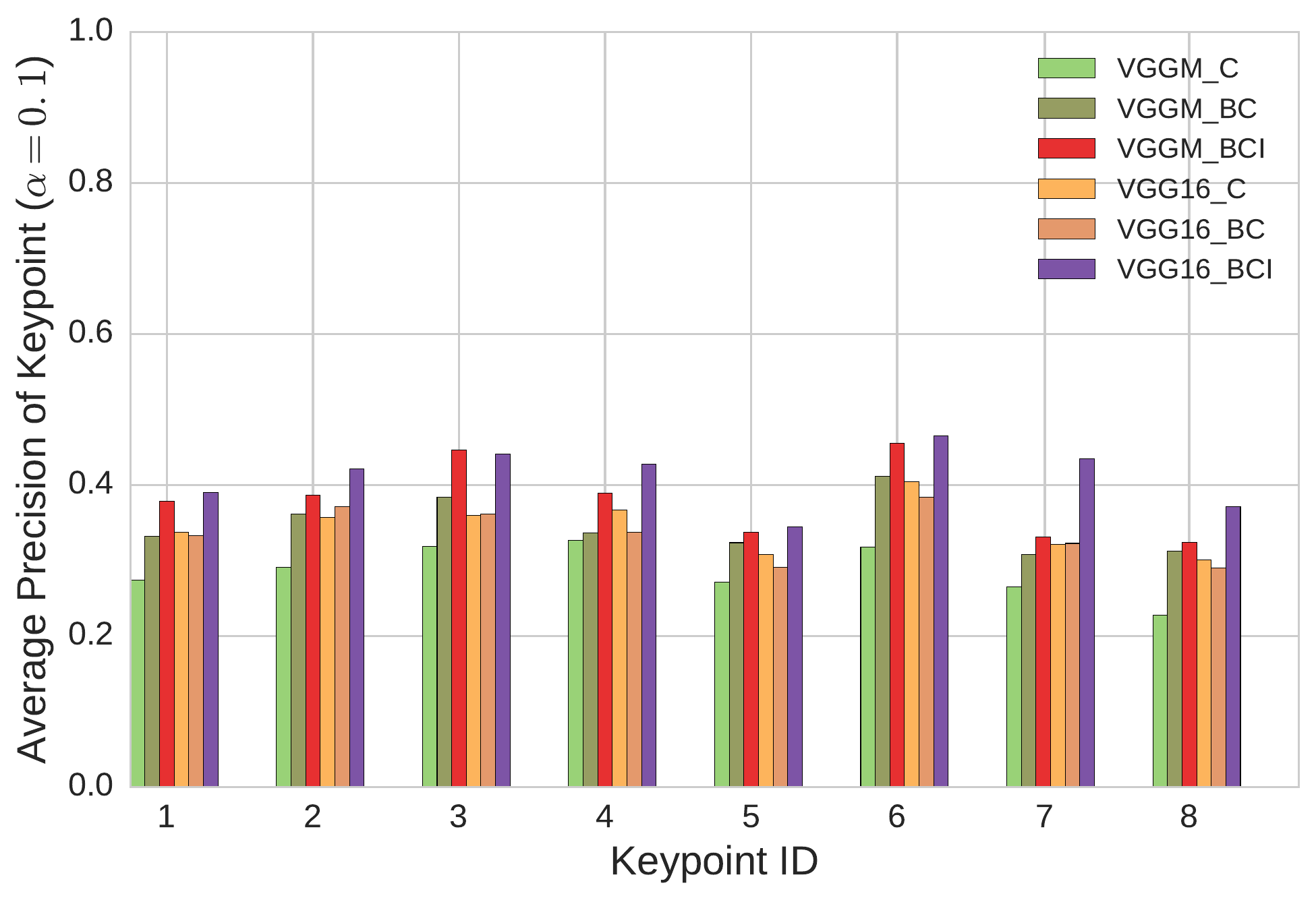} \\
(c) PCK vs. $\alpha$ & (d) Keypoint-wise APK   \\[6pt]
\includegraphics[width=0.5\textwidth]{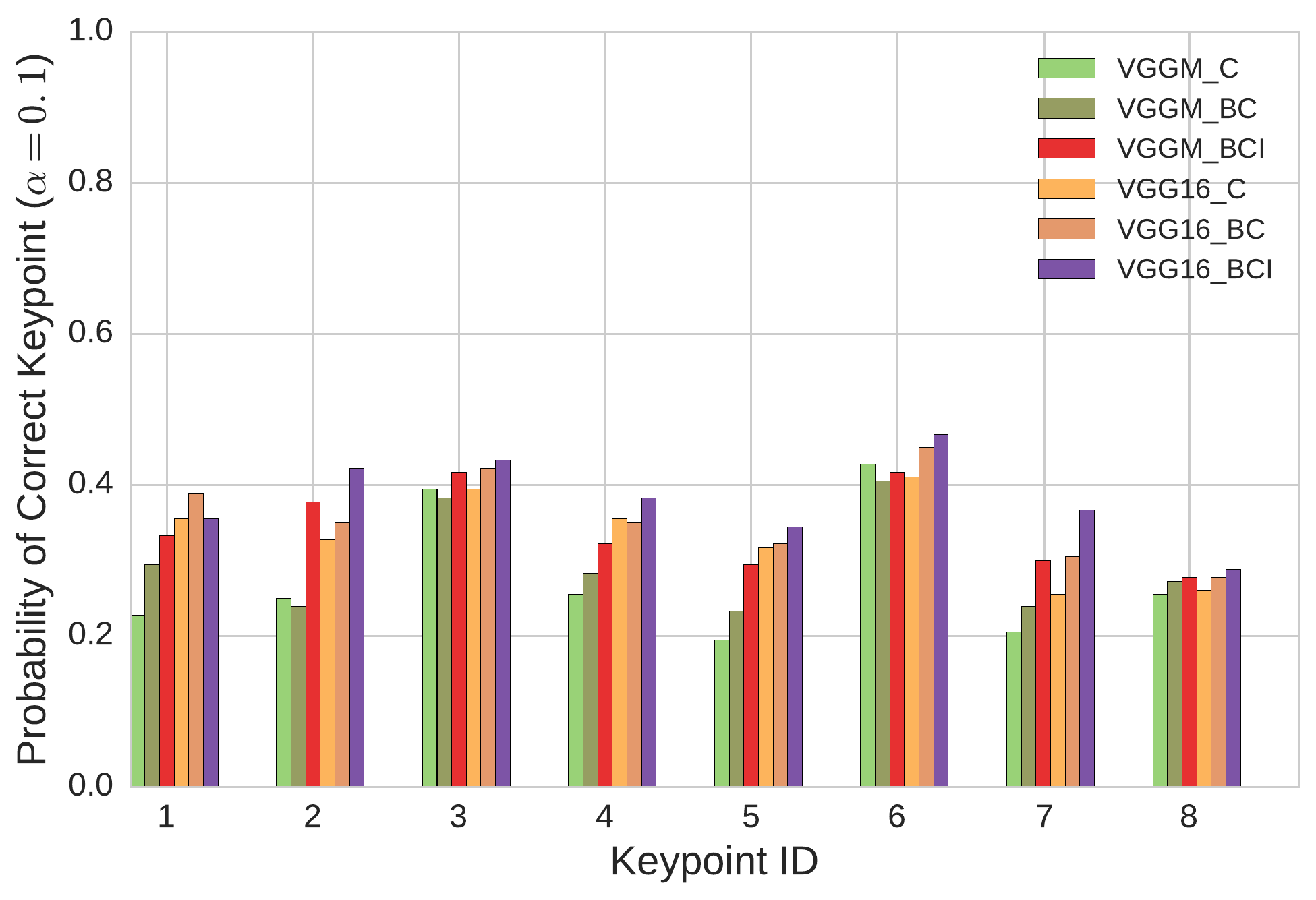}  &   \includegraphics[width=0.5\textwidth]{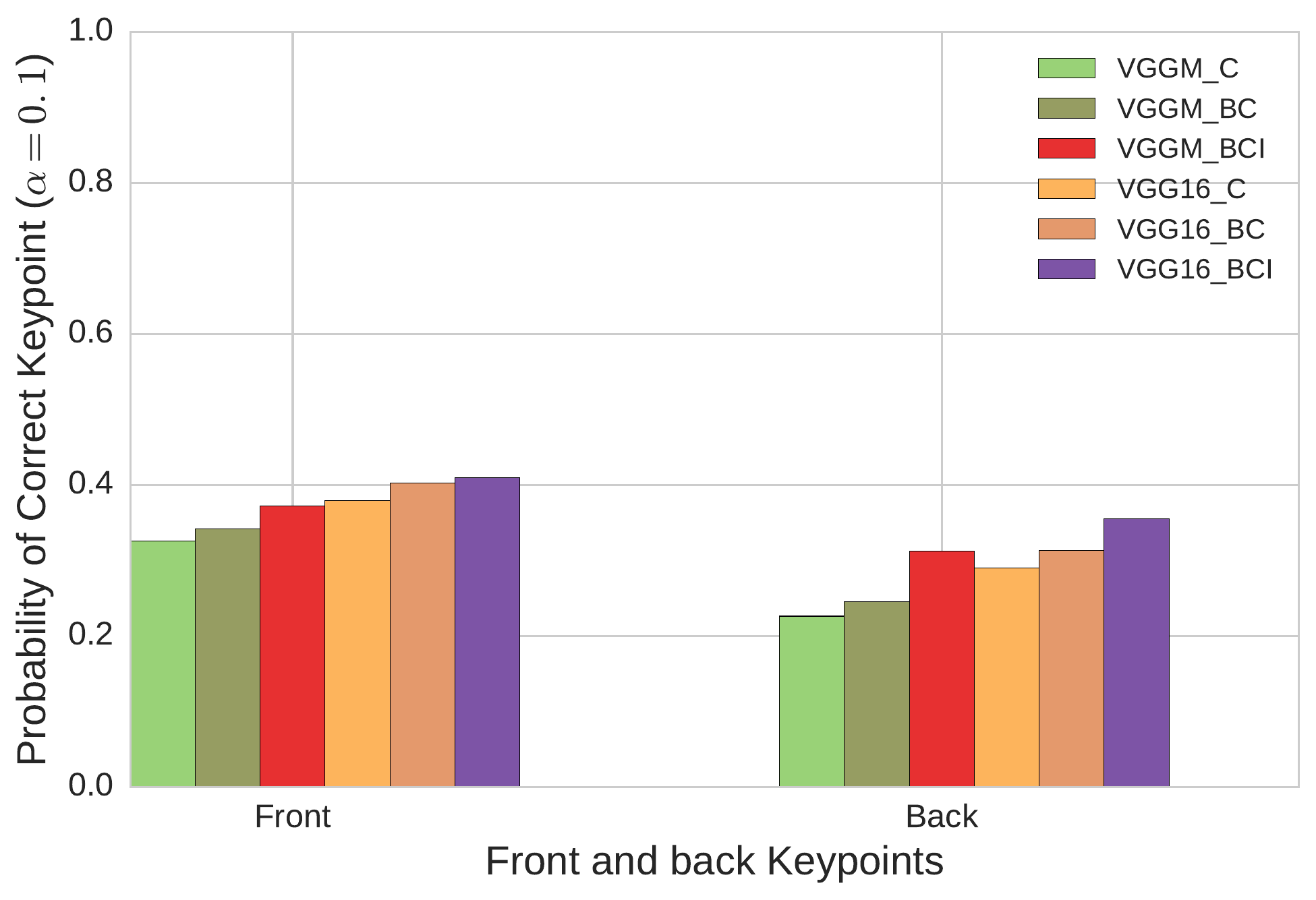} \\
(e) Keypoint-wise PCK & (f) Face-wise PCK \\[6pt]
\end{tabular}
\caption{{\bf Deep Cuboid Detector Evaluation Metrics.} APK: Average Precision of Keypoint, PCK: Probability of Correct Keypoint, $\alpha$: Normalized distance from GT corners, Order of keypoints: front-top-left, back-top-left, front-bottom-left, front-top-right, back-bottom-left, front-bottom-right, back-top-right, back-bottom-right. B = Bounding box loss, C = Corner loss, I = Iterative.}
\label{fig:evaluation-metrics}
\end{center}
\end{figure*}

We evaluate the system on two tasks: cuboid bounding box detection and cuboid keypoint localization. For detection, a bounding box is correct if the Intersection over Union (IoU) overlap is greater than $0.5$.\footnote{This is the standard way to evaluate a 2D object detector as was popularized by the PASCAL VOC object detection challenge~\cite{everingham2010pascal}}. Detections are sorted by confidence (the network's classifier softmax output) and we report the mean Average Precision (AP) as well as the entire Precision-Recall curve. For keypoint localization we use the Probability of Correct Keypoint (PCK) and Average Precision of Keypoint (APK) metrics~\cite{yang2013articulated}. PCK and APK are commonly used in the human pose estimation literature to measure the performance of systems predicting the location of human body parts like head, wrist, etc. PCK measures the  fraction of annotated instances that are correct when all the ground truth boxes are given as input to the system. A predicted keypoint is correct if its normalized distance from the annotation is less than a threshold ($\alpha$). APK, on the other hand, takes both detection confidence and keypoint localization into consideration. We use a normalized distance, $\alpha$, of $0.1$ which means that a predicted keypoint is considered to be correct if it lies within $0.1 \times max(height, width)$ pixels of the ground truth annotation of the keypoint. See Figure~\ref{fig:evaluation-metrics} to see these metrics reported on the SUN Primitive test set and samples of cuboid detections in Figure~\ref{fig:example-detections}.

For bounding box detection, our best network architecture achieves a mAP of $75.47$, which is significantly better than the HOG-based system of Xiao \etal.~\cite{xiao2012localizing} which reports a mAP of $24.0$.

\begin{table}
\begin{center}
\begin{tabular}{|l|c c c|}
\hline
Additional loss function & AP & APK & PCK \\
\hline\hline
 Bounding Box Loss  & 66.33 & - & -\\
 Corner Loss & 58.39 & 28.68 & 27.64 \\
 Bounding Box + Corner Loss & {\bf 67.11} & {\bf 34.62} & {\bf  29.38} \\
\hline
\end{tabular}
\end{center}
\caption{{\bf Multi-task learning Results} We first train a network using only the bounding box loss, then use the cuboid corner loss.}
\label{table:incremental}
\end{table}

\subsection{Multi-task learning}
\label{sec:multi-task}
We train multiple networks each of which performs different multiple tasks. We start off with a base network that just outputs bounding boxes around cuboids --- that is we train a network to perform general object detection using just the rectangle enclosing the cuboids. The above network outputs the class of the box and the bounding box regression values. Next, we train a different model with additional supervision about the location of the corners but this model does not output bounding box regression coordinates. Finally, we train a network that outputs both the bounding box regression values and the coordinates of the vertex. A corresponding term is added to the loss function for each additional task. From our experiments, we can see how adding more tasks affects the performance of the cuboid detector (see Table~\ref{table:incremental}).

\subsection{Iterative Feature Pooling}
\label{sec:ifp}

\begin{figure*}[ht]
\begin{center}
  \includegraphics[width=0.9\linewidth]{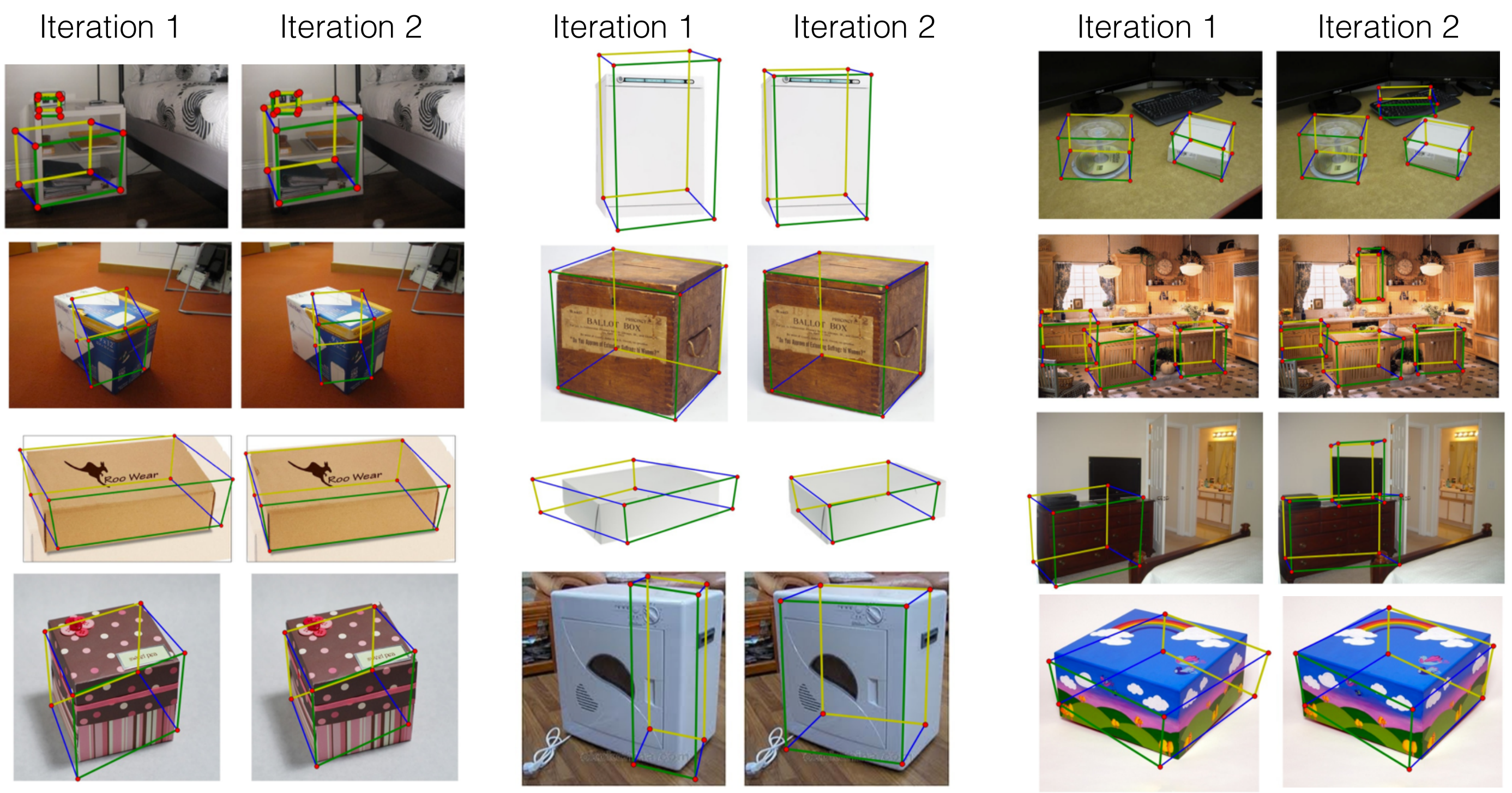}
\end{center}
  \caption{{\bf Vertex Refinement via Iterative Feature Pooling}. We refine cuboid detection regions by re-pooling features from \emph{conv5} using the predicted bounding boxes (see Section ~\ref{sec:ifp}).}
\label{fig:example-iterations}
\end{figure*}

In R-CNN, the final output is a classification score and the bounding box regression values for every region proposal. The bounding box regression allows us to move the region proposal around and scale it such that the final bounding box localizes just the object. This implies that our initial region from which we pooled features to make this prediction was not entirely correct. Hence, it makes sense to go back and pool features from the correct bounding box. We implement this in the network itself which means that we perform iterative bounding box regression while training and testing in exactly the same way. The input to the fully-connected layers is a fixed-size feature map that consists of the pooled features from different region proposals from \textit{conv5} layer. The R-CNN outputs are used for bounding box regression on the input object proposals to produce new proposals. Then features are pooled from these new proposals and passed through the fully-connected layers again. This model is now an ``any-time prediction system'' where for applications which are not bound by latency, bounding box regression can be performed more than once. Our results (see Table~\ref{table:ifp}) show that iterative feature pooling can greatly improve both bounding box detection and vertex localization (see Figure~\ref{fig:example-iterations}). There isn't a significant change in performance when we iteratively pool features more than twice.

\begin{table}
\begin{center}
\begin{tabular}{|l|c c c|}
\hline
Method & AP & APK & PCK \\
\hline\hline
 Corner Loss  & 58.39 & 28.68 & 27.64 \\
 Corner Loss + Iterative & 62.89& 33.98&35.56\\
 BB+Corner Losses & 67.11 & 34.62  & 29.38 \\
 BB+Corner Loss + Iterative & {\bf 71.72} &	{\bf 37.61} &  \textbf{36.53}\\
\hline
\end{tabular}
\end{center}
\caption{{\bf Results for Iterative Feature Pooling.} Our iterative algorithm improves the box detection AP by over 4\% and  PCK over 7\%.}
\label{table:ifp}
\end{table}

\subsection{Depth of Network}
We experiment with two base models: VGG16 and VGG-M. While VGG16 has a very deep architecture with $16$ layers, VGG-M is a smaller model with $7$ layers. We report results of these experiments in Table \ref{table:deeporshallow}. Interestingly, for this dataset and task, two iterations through the shallower network outperforms one iteration through the deeper network. Coupled with the fact the shallower network with iteration runs twice as fast, it can be a good choice for  deployment.

\begin{table}
\begin{center}
\begin{tabular}{|l|c c c c c|}
\hline
Method & AP & APK & PCK & Size & Speed \\
\hline\hline
 VGG-M  & 67.11 & 34.62 & 29.38 & \textbf{334MB} & \textbf{14fps}\\
  VGG-M +  I & 71.72 & 37.61 &  36.53& \textbf{334MB} & 10fps \\
 VGG16  & 70.50 & 33.65 & 35.83 & 522MB & 5fps \\

 VGG16 + I &  {\bf 75.47} &	{\bf 41.21} &	{\bf 38.27} & 522MB & 4fps\\
 
\hline
\end{tabular}
\end{center}
\caption{{\bf VGG-M (7 layer) vs. VGG16 (16 layer) base network.} I implies iterative feature pooling was performed. The deeper cuboid detector outperforms the shallower one.}
\label{table:deeporshallow}
\end{table}

\subsection{Effect of Training Set Size}
\label{sec:training-set-experiment}
We wanted to measure the impact of increasing the size of training data. We create three datasets of varying sizes: 1K, 2K and 3K images and train a common network (VGG-M + Iterative). Our results (see Table~\ref{table:data}) show significantly improved performance when using larger training set sizes.

\begin{table}
\begin{center}
\begin{tabular}{|c|c c c|}
\hline
Number of Images & AP & APK & PCK \\
\hline\hline
 1000  & 40.47 & 20.83  & 26.60\\
  2000  & 52.17 &	27.51 & 29.31 \\
 3000 & {\bf 71.72} & {\bf 37.61} & \textbf{36.53}\\
\hline
\end{tabular}
\end{center}
\caption{{\bf Performance vs. number of training images.} Deep cuboid detection benefits from more training images.}
\label{table:data}
\end{table}

\begin{figure}
\begin{center}
  \includegraphics[width=0.8\linewidth]{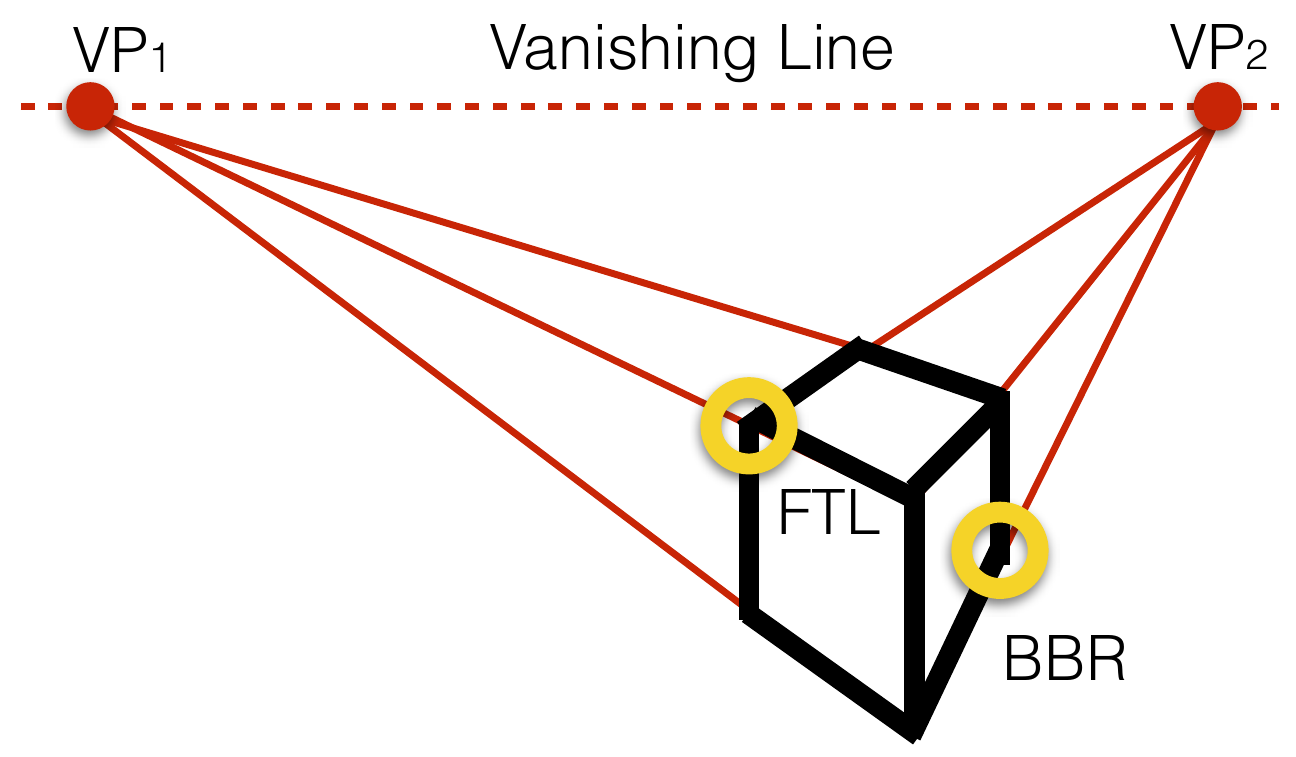}
\end{center}
  \caption{{\bf Cuboid Vanishing Points.} Vanishing points produced by extrapolating the edges of a cube can be used to reduce the number of parameters. We can drop the Front-Top-Left (FTL) and Back-Bottom-Right (BBR) vertices from our parametrization, and infer them using estimated VPs. }
\label{fig:vp}
\end{figure}

\subsection{Memory and Runtime Complexity}
Our cuboid detector is able to run at interactive rates on a Titan Z GPU while the HOG-based approach of Xiao \etal.~\cite{xiao2012localizing} takes minutes to process a single image. The real-time nature of the system is due to the fact that our implementation is based on top of Faster R-CNN. We anticipate that progress in object detection like SSD~\cite{liu2015ssd} can be leveraged to have a cuboid detector that is even faster. We also report the model size in MB (see Table~\ref{table:deeporshallow}). Further optimization can be done to reduce the size of the model such that it fits on mobile devices~\cite{rastegari2016xnor,iandola2016squeezenet,han2016deep}.

\begin{table*}
\begin{center}
\begin{tabular}{|l|c c c |  c c c|}
\hline
Method & AP & APK & PCK & PCK of BBR Corner & PCK of FTL Corner & PCK of Remaining Corners\\
\hline\hline
6 corners & 65.26 &	29.64 & 27.36	 &  24.44 & 21.11 & 28.89 \\
 8 corners  & {\bf 67.11} &	{\bf 34.62} &	{\bf 29.38}  & {\bf    27.22} & {\bf 29.44} & {\bf 29.73} \\
\hline
\end{tabular}
\end{center}
\caption{{\bf 8 corner vs. 6 corner parameterization} The 8 corner parameterization uses all of the cuboid's corners, whereas in the 6 corner version, the BBR and FTL corners are dropped (see Figure~\ref{fig:vp}) and inferred from the vanishing points. This shows how the network is able to do geometric reasoning and the over-parameterization adds robustness to the system (see Section~\ref{sec:parametrizations}). {\bf BBR:} Back-Bottom-Right and {\bf FTL:} Front-Top-Left}
\label{table:sixversuseight}
\end{table*}

\section{Discussion}
\label{sec:parametrizations}

So far, all of our networks output the cuboid's 8 vertices directly. However, a closer look at typical failures suggests that our parametrization might be to blame. Certain viewpoints have an inherent ambiguity (e.g., which face of the Rubik’s cube in Figure~\ref{fig:example-detections} should be labelled the front?).  Since our detector has trouble dealing with such configurations, in this section we explore and discuss alternate cuboid parametrizations. If we consider the world origin to coincide with camera center coordinates, the minimal parameterization of the cuboid can be done with $12$ numbers:

\begin{itemize}
    \item ($X$, $Y$, $Z$) - Centre of cuboid in 3D
    \item ($L$, $W$, $H$) - Dimensions of the cuboid
    \item ($\theta$, $\psi$, $\phi$) - 3 angles of rotation of the cuboid
    \item ($f$, $c_x$, $c_y$) - Intrinsic camera parameters
\end{itemize} 

For modern cameras, we can assume there is no skew in the camera and equal focal lengths. The over-parameterization we have in our network (as we predict 16 numbers) allows the network to produce outputs that do not represent cuboids. We experimented with several different re-parameterizations of a cuboid in an attempt to better utilize the geometric constraints. In general, our observation is that the network is able to learn features for tasks that have more visual evidence in the image and predict parameters which can be scaled properly for stable optimization. When dealing with 3D geometry and deep learning, proper parametrization is important. Even image-to-image transformations like homographies benefit from re-parametrization (\eg., the 4-point parametrization~\cite{detone2016deep}).

{\bf 6-corner parametrization:} We explore an alternate parameterization in which we predict only 6 coordinates and infer the locations of the remaining two coordinates using the fact that there are parallel edges in cuboids. The edges that are parallel in 3D meet at the vanishing point in the image. There are two pairs of parallel lines on the top and bottom face of the cuboid. These lines should meet at the same vanishing point~\cite{hartley2003multiple} as shown in Figure~\ref{fig:vp}. Using this fact, we can infer the position of the remaining two points. This allows us to parameterize our output in 12 numbers.

We carry out an experiment to be able to compare the 8 corner parameterization with the 6 corner parameterization. To do so, we decide to not use the ground truth data for two vertices while training. We leave one vertex each from the back and front faces (those whose detection rates (PCK) were the worst). We train a network to predict the location of the remaining 6 corners. We infer the location of the two dropped vertices using these 6 corners. We do so by first finding out the vanishing points corresponding to the 6 points we predicted. This reparameterization, however, leads to a reduction in performance (see Table~\ref{table:sixversuseight}). This degradation is due to the fact that we discard any visual evidence corresponding to the two inferred corners present in the image. Also, any error in prediction of one vertex due to occlusion or any other reason is directly propagated to the inferred corners. However, left to the CNN it learns multiple models to detect a cuboid. The network is free to use all visual evidence to localize the corners of the cuboid. It is interesting to note that a CNN is capable of doing pure geometric reasoning because in many cases the corner on the back does not have visual evidence in the image due to self-occlusion.

{\bf Vanishing point parametrization:} Another re-parameterization\footnote{The standard parameterization involves predicting the 8 vertices (16 parameters) of the cuboid directly.} uses locations of the two vanishing points and the slopes of six lines which will form the edges of the cuboid (see Figure~\ref{fig:vp}). Note that these vanishing points correspond to a particular cuboid and might be different from the vanishing point of the entire image. The intersection points of these 6 lines would gives us the vertices of the cuboid. However, the network fails to learn the location of the vanishing points as many a time they lie outside the region of interest and have little or confounding visual evidence in the region of interest or the entire image itself. It also becomes difficult to normalize the targets to predict the vanishing points directly. The slopes of the 6 lines can vary between $-\infty$ to $+\infty$. Instead of predicting the slope directly, we tried to regress to value of $sin(tan^{-1}(\theta))$. We found it difficult to train the network with this parameterization but there might exist a set of hyperparameters (loss weights, learning rates, solver, etc) for which we can train the above mentioned network.

\section{Conclusion}
In this paper we have presented an end-to-end deep learning system capable of detecting cuboids and localizing their vertices in RGB images of cluttered scenes. By casting the problem in a modern end-to-end deep learning framework, there is no need to design custom low-level detectors for line segments, vanishing points, junctions, etc. With Augmented Reality and Robotics applications in mind, we hope that the design and architecture of our system will inspire researchers to improve upon our results. Future work will focus on: training from much larger datasets and synthetic data, network optimization, and  various regularization techniques to improve generalization.

{\small
\bibliographystyle{ieee}
\bibliography{egbib}
}

\end{document}